%% file: arXiv.tex
\documentclass[]{article}
\usepackage{proceed2e}

\usepackage{times}

\usepackage{amsmath}  
\usepackage{graphicx} 
\usepackage{amssymb}
\usepackage{amsthm}

\newcommand{\bP}{\mathbb P}
\newcommand{\bb}[1]{\textbf{#1}}

\newcommand{\dd}{\downarrow}
\newcommand{\uu}{\uparrow}
\newcommand{\mc}[1]{\mathcal{#1}}
\newcommand{\VV}{\textit Val}

\usepackage{color}

\title{Multi-Context Models for Reasoning under Partial Knowledge:\\ Generative Process and Inference Grammar}

\author{ {\bf Ardavan S. Nobandegani} \\
Dept. of Electrical and Computer Engineering \\
McGill University\\
Montreal, QC H3A 0E9 \\
\And
{\bf Ioannis N. Psaromiligkos}  \\
Dept. of Electrical and Computer Engineering \\
McGill University\\
Montreal, QC H3A 0E9 \\
}

\begin{document}

\maketitle

\begin{abstract}
Arriving at the complete probabilistic knowledge of a domain, i.e., learning how all variables interact, is indeed a demanding task. In reality, settings often arise for which an individual merely possesses partial knowledge of the domain, and yet, is expected to give adequate answers to a variety of posed queries. That is, although precise answers to some queries, in principle, cannot be achieved, a range of plausible answers is attainable for each query given the available partial knowledge. In this paper, we propose the Multi-Context Model (MCM), a new graphical model to represent the state of partial knowledge as to a domain. MCM is a middle ground between Probabilistic Logic, Bayesian Logic, and Probabilistic Graphical Models. For this model we discuss: (i) the dynamics of constructing a contradiction-free MCM, i.e., to form partial beliefs regarding a domain in a gradual and probabilistically consistent way, and (ii) how to perform inference, i.e., to evaluate a probability of interest involving some variables of the domain.
\end{abstract}

\input{Intro}
\input{notation}
\input{MCM}
\input{Inference}
\input{Discussion}
\input{Appendix}

\newpage
\bibliographystyle{aaai}
\bibliography{ref}

\end{document}

%% file: Intro.tex
\section{INTRODUCTION}
At an abstract level, an individual (also referred to as a reasoner) is faced with a domain where by ``domain" we simply mean a collection of propositions or concepts which are mathematically encoded as Random Variables (RVs). To arrive at the complete probabilistic knowledge of the domain, i.e., to learn how all RVs in the domain probabilistically interact with one another, is indeed a demanding task. In reality, an individual is often faced with a domain for which she merely possesses \emph{partial} knowledge---that is, she  only knows how \emph{some} (not all) RVs in the domain interact. To make the setting under study more tangible, consider the following case. Suppose that the probabilistic knowledge of a domain is represented by a Probabilistic Graphical Model (PGM)  $\mathcal{B}$, e.g., a Bayesian Network (BN). Then the reasoner comes across a new RV, say $\boldsymbol\psi$, and would like to incorporate it into $\mathcal{B}$ so as to achieve the complete probabilistic knowledge of the new domain (which now also includes $\boldsymbol\psi$). However, incorporation of $\boldsymbol\psi$ into $\mathcal{B}$ would require knowledge of how $\boldsymbol{\psi}$ is probabilistically related to all the RVs already present in $\mathcal{B}$; a knowledge which may be, quite plausibly, unavailable to the reasoner. An interesting question that now arises is how to handle  situations where only partial knowledge as to how $\boldsymbol\psi$ is probabilistically related to $\mathcal{B}$ is available. An example would be when the reasoner merely knows how $\boldsymbol\psi$ interacts probabilistically with only one RV, say $\boldsymbol{\phi}$, in $\mathcal{B}$. 

In this paper, a graphical model, namely, the Multi-Context Model (MCM) is proposed to represent the setting in which only partial probabilistic knowledge of a domain is available to the reasoner. More specifically, MCM is a graphical language to represent  settings in which the Joint Probability Distribution (JPD) over all RVs is not available, but what is available instead is the JPDs over a collection of subsets of RVs of the domain (referred to as sub-domains or \emph{contexts}). These contexts are potentially overlapping, i.e., they could share some RVs. As pointed out elegantly in \cite{pearl1990reasoning}, \emph{``this state of partial knowledge is more common, because we often begin thinking about a problem through isolated frames, paying no attention to interdependencies."} Along the same line of thought, it is  plausible to assume that the probabilistic knowledge of the domain at the early primitive stage consists of a collection of disjoint contexts and as the reasoner acquires more knowledge as to how the variables in the model are related to one another and thus probabilistically interact, contexts gradually go through a process very much like an evolution: contexts start to share some variables, overlaps begin to emerge and, once enough knowledge is obtained, a number of contexts could merge thereby giving rise to bigger contexts. This naturally raises the following fundamental question: How could a collection of consistent, probabilistically sound, and potentially overlapping contexts emerge \emph{gradually} over the course of time? In an attempt to answer this question we present a generative process of constructing a contradiction-free MCM. Finally, we would like to note that the special case where the whole domain is modeled as a single context corresponds to the conventional way of modeling the probabilistic knowledge of a domain using a single PGM, e.g., by some BN.

Another yet crucial question which we  address in this work---which is another motivation behind the development of the MCM---is how the task of inference (i.e., the evaluation of some probability of interest which is hereafter referred to as \emph{query}) should be carried out in a domain which is modeled according to some MCM. A query does not necessarily belong to any one of the contexts in particular and, in fact, may involve RVs from different contexts. 

The paper is structured as follows. After introducing the notation in Sec. \ref{sec:term},  we define in Sec. \ref{sec:gen_p} the MCM and drawing on the notion of probabilistic conditioning, a generative process of constructing a contradiction-free MCM is discussed. Then, in Sec. \ref{sec:inference} we elaborate on the problem of inference in a multi-context setting, i.e., in a domain whose probabilistic knowledge is encoded as an MCM. In Sec. \ref{sec:work} we discuss the relevant past work and comment on the proposed model. Finally, Sec. \ref{sec:conclusion} concludes the paper.

%% file: notation.tex
\section{TERMINOLOGY AND NOTATION}
\label{sec:term}
In this section we present the mathematical notation and the terminology employed in this paper. 
Random quantities are denoted by bold-faced letters; their realizations are denoted by the same letter but non-bold. More specifically, RVs are denoted by lower-case bold-faced letters, e.g., $\bb{x}$, while random vectors are denoted by upper-case bold letters, e.g., $\bb{X}$. $\textit{Val}(\cdot)$ denotes the set of values a random quantity can take, e.g., $\textit{Val}(\bb{x})$ is the set of all possible realizations of the RV $\bb{x}$. In this paper, we assume that all random quantities are discrete.

The JPD over the RVs $\bb{x}_1,\cdots,\bb{x}_n$ is denoted by $\bP(\bb{x}_1,\cdots,\bb{x}_n)$; when $\bb{x}_1,\cdots,\bb{x}_n$ comprise a vector $\bb{X}$ then $\bP(\bb{X}):=\bP(\bb{x}_1,\cdots,\bb{x}_n)$. We will use the notation $\bb{x}_{1:n}$ to denote the sequence of $n$ RVs $\bb{x}_1,\cdots,\bb{x}_n$. To simplify presentation and to prevent our expressions from becoming cumbersome, we incur the following abuse of notation: We denote the  probability $\bP(\bb{x}=x)$ by $\bP(x)$ for some RV $\bb{x}$ and its realization $x\in \textit{Val}(\bb{x})$. Also, $\bP(\bar{x}):=\bP(\bb{x}\neq x)=1-\bP(x)$ for some $x\in \textit{Val}(\bb{x})$, i.e., $\bP(\bar{x})$ is the probability that $\bb{x}$ takes on any value other than $x$. For conditional probabilities we will use the notation $\bP(x|y)$ instead of $\bP(\bb{x}={x}|\bb{y}={y})$.  Similar notations will be used for the case of random vectors, i.e., $\bP(X):=\bP(\bb{X}=X)$, $\bP(\bar{X}):=\bP(\bb{X}\neq X)=1-\bP(\bb{X}=X)=1-\bP(X)$, and $\bP(X|Y):=\bP(\bb{X}={X}|\bb{Y}={Y})$.

The subscript $\downarrow$ on a probability, e.g.,  $\bP(x|y)_\dd$, denotes the minimum value the probability can take subject to the constraints induced by the available probabilistic knowledge. Likewise,  the subscript $\uparrow$ on a probability denotes the maximum value the probability can take. Finally, the operator $[\cdot]^+$ gives the positive part of its argument, i.e., $[a]^+:=\max\{0,a\}$ for any real-valued $a$.

%% file: MCM.tex
\section{MULTI-CONTEXT MODEL}
\label{sec:gen_p}
\begin{figure}[h!] 
  \centering
    \includegraphics[width=0.3 \textwidth]{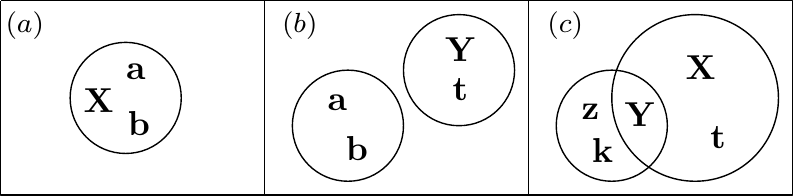}
  \caption{Graphical representation of contexts: (a) Context associated to $\bP(\bb{a},\bb{b},\bb{X})$. (b) Two disjoint contexts associated to $\bP(\bb{a},\bb{b})$ and $\bP(\bb{Y},\bb{t})$. (c) Two overlapping contexts associated to $\bP(\bb{X},\bb{Y},\bb{t})$ and $\bP(\bb{Y},\bb{z},\bb{k})$. The random vector $\bb{Y}$ is referred to as the \emph{induced} part in Sec. \ref{sec:gen_p}.}
  \label{fig_termin}
\end{figure}

As explained earlier, a \textit{domain} is simply the set of all Random Variables (RVs) at hand. A \textit{context} comprises a  collection of RVs for which their JPD is precisely known, see Fig. \ref{fig_termin}(a). In general, two contexts could be disjoint (Fig. \ref{fig_termin}(b)) or overlapping (Fig. \ref{fig_termin}(c)).

A \textit{Multi-Context Model (MCM)} encodes the probabilistic knowledge of a domain as a collection of possibly overlapping contexts. This enables the handling of situations in which comprehensive knowledge of a domain is not available, but partial information is, in the form of  JPDs of some subsets of the domain. Let us first motivate the proposed MCM by entertaining a simple yet enlightening example.

\subsection{MOTIVATING EXAMPLE}
\begin{figure}[h!] 
  \centering
    \includegraphics[width=0.13\textwidth]{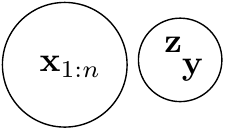}
  \caption{Problem statement as an MCM.}
  \label{fig_motivation_toy_problem_1}
\end{figure}

Consider a domain consisting of the RVs $\bb{y},\bb{z}$ in addition to a set of $n$ RVs, $\bb{x}_{1:n}$. A reasoner has formed a partial belief as to the probabilistic connections between the variables of the domain. More specifically, the reasoner knows precisely the JPDs $\bP(\bb{y},\bb{z})$ and  $\bP(\bb{x}_{1:n})$ but not the JPD $\bP(\bb{y},\bb{z}, \bb{x}_{1:n})$. This setting is described by an MCM that consists of two disjoint contexts, one associated to RVs $\bb{y},\bb{z}$ and the other to $\bb{x}_{1:n}$,  as shown in Fig. \ref{fig_motivation_toy_problem_1}.

Assume that the following query is posed: Given the available information, what could be said about $\bP(y|x_i)$ for some $i=1,\cdots,n$? The RVs $\bb{y}$ and $\bb{x}_i$ belong to different contexts, therefore, the JPD of $\bb{y}$ and $\bb{x}_i$, $\bP(\bb{x}_i,\bb{y})$, is not available. The best one can hope for is to derive the range within which $\bP(y|x_i)$ varies, namely, $[\bP(y|x_i)_\dd,\bP(y|x_i)_\uu]$. Let us for the moment assume the objective is to find $\bP(y|x_i)_\dd$.
Based on the conventional methodology, i.e., the approach adopted by past work (cf. \cite{andersen1990probabilistic,andersen1994bayesian,hansen1995models} and references therein)  one has to write down \emph{all} the information as a list of linear equations and solve it as a Linear Program (LP). The main drawback of the conventional approach is that it cannot distinguish between what information is relevant and what is irrelevant for the posed query, and hence what needs to and what need not be considered in answering the query. The price for this is that the number of parameters required to merely  formulate the query as an LP is exponential in $n$. 

The key point, however, is that what information is relevant (or irrelevant) depends directly on the posed query, i.e., it is query-dependent. The main advantage of  the proposed MCM over previous approaches is that it enables answering a query in a computationally efficient manner by distinguishing the relevant information from the irrelevant for the given query. This is realized thorough adopting the notion of \emph{inference grammar}; a concept which will be systematically defined later. For our example, following the inference rule we will provide in Sec. \ref{Sec:Grammar}, one can easily get $\bP(y|x_i)_\dd=[\frac{\bP(y)-\bP(\bar{x}_i)}{\bP(x_i)}]^+$.

The task of inference in an MCM is carried out on two different levels, which makes the task more computationally efficient: 
\begin{itemize}
\item[(i)] High-Level Reasoning: at this level, through the use of inference grammar, the relevant quantities are identified (e.g., $\bP(y)$ and $\bP(\bar{x}_i)$ in the case of our example). 
\item[(ii)] Low-Level Reasoning: the relevant quantities, identified in (i), can then be computed by employing inference algorithms which take advantage of the potentially rich independence structure governing the contexts. For example, it could very well be the case that for the JPD associated to $\bb{x}_{1:n}$ a large number of conditional independence relations hold. In that case, stating the derivation of $\bP(\bar{x}_i)$ (i.e., $1-\bP(x_i)$) as an LP would be  computationally inefficient\footnote{The number of parameters required just to state the problem as an LP is exponential in $n$.} but unnecessary. Indeed, the task of finding $\bP(\bar{x}_i)$ could be accomplished in a computationally efficient way using one of the many inference methods developed for probabilistic graphical models; a key point that the previous approaches do not take advantage of. 
\end{itemize}
As a final step, in order to derive the lower/upper bound to the posed query, the quantities identified in (i) and subsequently calculated in (ii) are stated and solved as an LP. 

The idea behind ``high-level reasoning" will be explained and clarified further in Sec. \ref{Sec:Grammar} and \ref{sec:transformation}, while the concept of ``low-level reasoning" will be discussed in Sec. \ref{Intra_Contextual_Inference_Problem}.

\subsection{GENERATIVE PROCESS OF CONTRADICTION-FREE MCMS} 
\label{Sec:Gen}
The objective of the generative process we describe in this section is to provide a way to consistently\footnote{That is, without introducing any form of contradictory result with respect to any probability assignment.} construct contexts, in a sequential manner, over a set of RVs. The act of constructing a context, i.e., of assigning a JPD to a subset of RVs, corresponds to forming a \emph{subjective}\footnote{One must not interpret the subjectivity of belief as ``total disconnectivity from the reality." Thus, we adopt the Bayesian interpretation of probability in this section. The avid reader is referred to \cite{chalmers2013thing}. An adherent to the frequentist interpretation of probability could think of contexts as being empirically constructed from a collection of data  and thus skip Sec. \ref{Sec:Gen} and proceed directly to the next section.} belief over those RVs. In this light, the act of constructing multiple contexts corresponds to \emph{gradually} forming subjective beliefs over a number of subsets of variables in the domain; hence every context symbolizes an established belief over the RVs involved in that context.

\begin{figure}[h!] 
  \centering
    \includegraphics[width=0.31\textwidth]{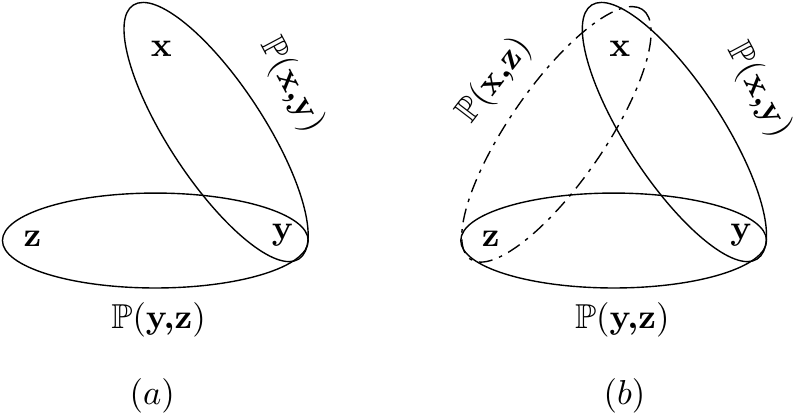}
  \caption{Generative process for contradiction-free Multi-Context Model.  The dash-dotted contexts cannot be freely assigned.}
  \label{fig_consis_3var_2pic}
\end{figure}

We introduce this problem by considering a simple case shown in Fig. \ref{fig_consis_3var_2pic}(a).
Suppose there are three RVs, namely, $\bb{x},\bb{y},$ and $\bb{z}$, present in the domain and let us consider the following question: Could one assign $\bP(\bb{x},\bb{y})$ and $\bP(\bb{y},\bb{z}),$ \textit{freely} and \emph{gradually} in a consistent manner, over the three variables without introducing any sort of contradiction? It is easy to verify that the answer is positive. Indeed, one could start off by assigning $\bP(\bb{x},\bb{y})$.
This assignment would, of course, induce the marginal $\bP(\bb{y})$ and one can write $\bP(\bb{y},\bb{z})=\bP(\bb{y})\bP(\bb{z}|\bb{y})$. Then, to complete this task, one would just need to proceed with assigning $\bP(\bb{z}|\bb{y})$. This process could be referred to as a \textit{generative} process of the assignment of $\bP(\bb{x},\bb{y})$ and $\bP(\bb{y},\bb{z})$ over $\bb{x},\bb{y}$, and $\bb{z}$ without introducing any inconsistencies, in a gradual manner. Indeed, free-assignment refers to the act of freely assigning the non-induced, e.g., $P(\bb{z}|\bb{y})$, part of the \emph{to-be-formed} belief, e.g., $P(\bb{y},\bb{z})$. In other words, free-assignment signifies the observation that the already-formed belief does not impose any constraints on the non-induced part of the to-be-formed belief.

Let us now consider the case shown in Fig. \ref{fig_consis_3var_2pic}(b). 
Could one assign $\bP(\bb{x},\bb{y}),\bP(\bb{y},\bb{z}),$ and $\bP(\bb{x},\bb{z})$ freely and {gradually} in a consistent manner over the three variables without introducing any sort of contradiction?
After some investigation, one can see that the answer is negative \cite{pearl1985bayesian}.  Not surprisingly, the reason  for this has to do with the existence of a loop in the model: once $\bP(\bb{x},\bb{y})$ and $\bP(\bb{y},\bb{z})=\bP(\bb{y})\bP(\bb{z}|\bb{y})$ are assigned\footnote{$\bP(\bb{y})$ is induced by the assignment of $\bP(\bb{x},\bb{y})$.}, then $\bP(\bb{x},\bb{z})$ cannot be assigned freely. This is due to the fact that $\bP(\bb{x},\bb{z})$ has to satisfy some non-trivial conditions imposed by the already assigned contexts $\bP(\bb{x},\bb{y})$ and $\bP(\bb{y},\bb{z})$ \cite{pearl1985bayesian}.

In summary, whenever it comes to generating a new context, the JPD associated to that context has to be separated into two parts: (i) the part induced by the already existing contexts, and (ii) the part containing new variables which have never been so far associated to any context (i.e., non-induced part). The key point in the generation of contradiction-free MCMs is that the former part has to be induced by some context which, itself, is already present in the domain. That is, all the induced parts have to be already contained within some context. Otherwise, to include the induced parts---each constrained by the context it is already in---in a new context, the newly created context would have to satisfy some nontrivial constraints and therefore could not be \emph{freely} assigned.

\begin{figure}[h!] 
  \centering
    \includegraphics[width=0.16\textwidth]{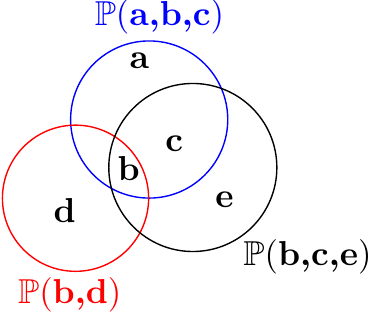}
  \caption{MCM for $\bP(\bb{a},\bb{b},\bb{c}), \bP(\bb{b},\bb{d})$, and $\bP(\bb{b},\bb{c},\bb{e})$.}
  \label{fig_mcx_1}
\end{figure}
Let us discuss one final case to further clarify the process.
Consider the multi-context model in Fig. \ref{fig_mcx_1}. Could this model be constructed freely and gradually in a probabilistically consistent manner? The answer is positive. We first assign $\bP(\bb{a},\bb{b},\bb{c})$, then we assign $\bP(\bb{b},\bb{c},\bb{e})=\bP(\bb{b},\bb{c})\bP(\bb{e}|\bb{b},\bb{c})$ where $\bP(\bb{b},\bb{c})$ is induced by our first assignment of $\bP(\bb{a},\bb{b},\bb{c})$. Finally, we assign $\bP(\bb{b},\bb{d})=\bP(\bb{b})\bP(\bb{d}|\bb{b})$ where $\bP(\bb{b})$ is induced by our first assignment of $\bP(\bb{a},\bb{b},\bb{c})$. A closer look reveals that this is not the only way we can  gradually construct a contradiction-free model in this case: we could have performed the assignments in a different order\footnote{Yet, this is not always the case: suppose there are four RVs in the domain, namely, $\bb{a},\bb{b},\bb{c}$ and $\bb{d}$ and we would like to assign $\bP(\bb{a},\bb{b}),\bP(\bb{b},\bb{c}),$ and $\bP(\bb{c},\bb{d})$. Performing the assignments in the order $(1)-\bP(\bb{a},\bb{b}),(2)-\bP(\bb{b},\bb{c}), (3)-\bP(\bb{c},\bb{d})$ would not introduce any inconsistencies, in contrast to using the order $(1)-\bP(\bb{a},\bb{b}),(2)-\bP(\bb{c},\bb{d}),(3)-\bP(\bb{b},\bb{c})$.}. Of course, the only thing which would have been different would be the induced probabilities. That is, if one does the assignment in the following order: (1)$-\bP(\bb{b},\bb{d})$, (2)$-\bP(\bb{a},\bb{b},\bb{c})$, (3)$-\bP(\bb{b},\bb{c},\bb{e})$ then the first assignment of $\bP(\bb{b},\bb{d})$ will induce $\bP(\bb{b})$ for the second assignment of $\bP(\bb{a},\bb{b},\bb{c})=\bP(\bb{b})\bP(\bb{a},\bb{c}|\bb{b})$ and the second assignment will induce $\bP(\bb{b},\bb{c})$ for the third assignment $\bP(\bb{b},\bb{c},\bb{e})=\bP(\bb{b},\bb{c})\bP(\bb{e}|\bb{b},\bb{c})$.

%% file: Inference.tex
\section{INFERENCE IN MCMS}
\label{sec:inference}
In this section we consider \emph{evidential} inference problems in multi-context settings. The objective is to evaluate (to the extent possible) a probability of the form $\bP(\bb{O}=O|\bb{E}=E)$, called a \emph{query}, where $\bb{O}$ and $\bb{E}$ are two mutually exclusive sets of RVs.  The set $\bb{E}$ is the set of evidence variables and $\bb{O}$ is the set of RVs for which we are interested in knowing with what probability they take on the value $O$, upon the observation of $\bb{E}=E$. In multi-context settings, inference problems can be categorized into two broad classes: 
\begin{itemize}
\item Intra-Contextual Inference Problems: For which the sets $\bb{E}$ and $\bb{O}$ both belong to the same context. \item Inter-Contextual Inference Problems: For which the sets $\bb{E}$ and $\bb{O}$ do not belong to a single context and, therefore, more than one context is involved in the inference problem.
\end{itemize}
In what follows, we will elaborate on these two cases.

\subsection{INTRA-CONTEXTUAL INFERENCE PROBLEM}
\label{Intra_Contextual_Inference_Problem}
One advantage of MCMs is that, once an inference problem is found to be an intra-contextual inference problem, one can take advantage of the rich independence structure potentially governing the context to accomplish the task of inference in a computationally efficient way. For instance, if the probabilistic knowledge of a context is presented in a form of a BN, then one can benefit from a variety of exact or approximate methods already developed for BNs. For a comprehensive study of such methods the reader is referred to \cite{koller2009probabilistic}. Hence, it is of great interest to have contexts whose probabilistic knowledge can be represented in some form of a PGM with sufficiently rich independence structure for which inference problems can be solved in a computationally efficient way. For example, if the probabilistic knowledge of a context is to be modeled according to some BN, we would like that BN to be as sparsely connected as possible and enjoy low tree-width to ensure computational efficiency for the task of inference \cite{chandrasekaran2012complexity}.

\subsection{INTER-CONTEXTUAL INFERENCE PROBLEM: INFERENCE GRAMMAR}
\label{Sec:Grammar}
In this section, we turn our attention to the task of inter-contextual inference. The RVs involved in the query for the inter-contextual inference problem do not belong to a single context. For this reason, the answer to the query is inevitably in the form of an interval indicating a lower and upper bound for the query. Since $\bP(E|O)+\bP(\bar{E}|O)=1$ we have $\bP(E|O)_\uparrow=1-\bP(\bar{E}|O)_\dd$. Therefore, we can focus our attention on the minimization problem (i.e., identifying a lower bound to the probability of interest) realizing that any maximization problem (i.e., identifying an upper bound to the probability of nterest) could be cast as a minimization problem and vice versa.

\begin{figure}[h!] 
  \centering
     \includegraphics[width=0.47\textwidth]{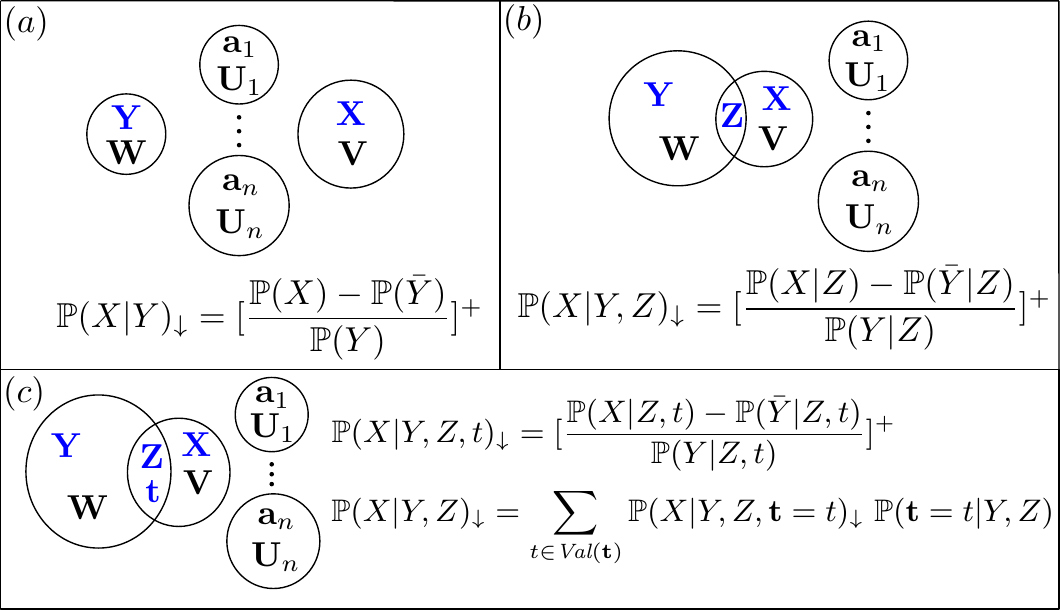}
  \caption{Sample inference rules given for some inter-contextual inference problems. The RVs involved in the query are shown in blue.}
  \label{fig_grammar_cases}
\end{figure}

First, we are going to consider some simple queries which are posed to some example MCMs. These MCMs are depicted in Fig. \ref{fig_grammar_cases}(a-c). The goal here is to develop some insight as to which variables are indeed relevant and which are deemed irrelevant for a given query and the corresponding MCM. 

We begin by considering a simple case: the disjoint MCM shown in Fig. \ref{fig_grammar_cases}(a). The rule to evaluate $\bP(X|Y)_\dd$ is also given in Fig. \ref{fig_grammar_cases}(a). Interestingly enough, the expression only requires the intra-contextual quantities $\bP(X)$ and $\bP(Y)$ and it does not depend on any other RV present in the domain. In other words, as far as $\bP(X|Y)_\dd$ is concerned, the MCM shown in Fig. \ref{fig_grammar_cases}(a) is equivalent to a much simpler MCM: the one corresponding to having only two disjoint contexts described by $\bP(\bb{X})$ and $\bP(\bb{Y})$. Next, we take the MCM given in Fig. \ref{fig_grammar_cases}(b) where there is an overlap between the context containing $\bb{X}$ and the one containing $\bb{Y}$. The overlapping part consists of the random vector $\bb{Z}$. The rule to evaluate $\bP(X|Y,Z)_\dd$ is given in Fig. \ref{fig_grammar_cases}(b). Now, consider the MCM shown in Fig. \ref{fig_grammar_cases}(c) where we have the same setting we had in previous case but a new random variable $\bb{t}$ is added in the overlapping region. Notice that the expression for $\bP(X|Y,Z,t)_\dd$ given in Fig. \ref{fig_grammar_cases}(c) is the same expression given for $\bP(X|Y,Z)_\dd$ in Fig. \ref{fig_grammar_cases}(b) with the substitution of $Z,t$ instead of $Z$. That is, $Z$ in Fig. \ref{fig_grammar_cases}(b) and $Z,t$ in Fig. \ref{fig_grammar_cases}(c) are representing the same thing, namely, ``all the variables in the overlapping region", and in that respect, they are ultimately the same. The rules are very much like sentences in predicate logic for which variables merely serve as place-holders.

The derivation of the rules given in Fig. \ref{fig_grammar_cases}(a-c) is not presented here. However, using the proof presented in Sec. A-II of Appendix (to identify the relevant variables) and subsequently following the methodology outlined in Sec. A-III of  Appendix (to visualize the partitions and reason out the extent they overlap) it should be straightforward to derive the presented rules. 

The sample set of rules presented is by no means exhaustive, nonetheless, due to the idea of {context transformation that will be} discussed in Sec. \ref{sec:transformation}, they can be applied to a wide range of interesting inter-contextual inference problems. We would like to clarify that our ultimate objective is \emph{not} to compute and provide the complete set of rules { that can answer all possible queries and for all possible MCMs}, since simply, the set is infinite in size. What we need, therefore, is an algorithm, let us call it  $\mathcal{I}^\ast$, that can provide the answer to the posed query being given an MCM as an input. The presented rules provide insights and hints to the nature of $\mathcal{I}^\ast$ which needs to be devised to ideally handle \emph{any} arbitrary query posed to \emph{any}\footnote{Although we believe that the MCMs generated through the generative process outlined in Sec. \ref{Sec:Gen} are more cognitively plausible, nonetheless, from a pure mathematical point of view, it would be of interest to find an algorithm which could handle \emph{any} MCM.} MCM. In a sense, we can get a glimpse of the nature of $\mathcal{I}^\ast$  through analyzing the presented rules. In other words, the derived rules serve as a lens through which one can study $\mathcal{I}^\ast$. In Sec. A-I of Appendix  a simple version of $\mathcal{I}^\ast$ that can handle arbitrary MCMs is outlined.

The motivation behind giving this sample set of rules can now be summarized in the following. 
\begin{enumerate}
\item To shed light on the general nature of a rule (which reflects on the nature of $\mathcal{I}^\ast$). More specifically, to illustrate that a rule enjoys two key properties, namely: (i) scale-invariance, (ii) resemblance to sentences in predicate logic, in that in both cases, variables are mere place-holders. For this resemblance we refer to $\mathcal{I}^\ast$ as \emph{inference grammar}.

\item To demonstrate that a rule is telling us which intra-contextual quantities are essential and which are irrelevant for a particular inter-contextual query.

\item To emphasize the key property that a rule derived under a specific MCM remains valid for and can be applied to infinitely many other MCMs all of which are linked through the notions of nestedness and transformation; hence generalization is achieved.

\item To lay down the foundation of \emph{transformation} and \emph{nestedness} which both play crucial roles in understanding the underlying machinery behind $\mathcal{I}^\ast$.
\end{enumerate}  
Next, we discuss another key property of the inference rules, namely, that of \emph{scale-invariance}. Consider once again the case in Fig. \ref{fig_motivation_toy_problem_1}. Now let us derive $\bP(x_i|y)_\dd,$ and $\bP(X|y)_\dd$ where $\bb{X}\triangleq \bb{x}_{1:n}$. Using the rule given in Fig. \ref{fig_grammar_cases}(a), one arrives at the following results: $\bP(x_i|y)_\dd=[\frac{\bP(x_i)-\bP(\bar{y})}{\bP(y)}]^+,$ and $\bP(X|y)_\dd=[\frac{\bP(X)-\bP(\bar{y})}{\bP(y)}]^+$.
In other words, the expressions remain the same, regardless of the dimension of the quantity of interest, i.e., be it a single RV or be it a random vector comprised of many RVs. In this respect, once again, the inference rules resemble expressions in predicate logic. The intuition on the scale invariance is provided in Sec. A-III of Appendix.

It is worth noting that $\mathcal{I}^\ast$ formulates the inter-contextual inference problem as a Linear Programming (LP) optimization (cf. Sec. A-I of Appendix). The key issues to consider are: (i) what RVs have to be included in the LP, and (ii) the abstraction level $\mathcal{I}^\ast$ should choose to encode the RVs identified in step (i) for the LP, i.e., the parametrization of RVs identified in step (i) for the LP. In what follows, the concepts of nestedness and transformation are put forth. Once the two are introduced, one could apply a single rule (e.g., one in Fig. \ref{fig_grammar_cases}(a)) to a much larger number of MCMs; in fact to infinitely many MCMs.

\subsection{INTER-CONTEXTUAL INFERENCE PROBLEM: NESTEDNESS AND TRANSFORMATION}
\label{sec:transformation}
\begin{figure}[h!]
  \centering
  \includegraphics[width=0.49\textwidth]{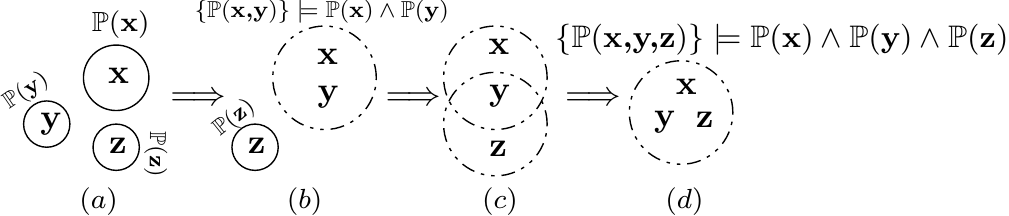}
  \caption{Inter-Contextual Inference Problem: Transformation and hierarchical construct. As one proceeds from the left to the right, a more comprehensive knowledge of domain is assumed to be available, of course hypothetically.}
  \label{fig_transform_1}
\end{figure}
The nested property, or \emph{nestedness}, refers to the fact that every MCM can be considered as an element of a family of MCMs. That family contains all MCMs which through marginalization can produce the original MCM. In such a case we simply say that the nested property holds between the original MCM and the family. The process of going from the original MCM to one of the members of the family is referred to as \emph{transformation}. For example, the MCM containing three contexts $\{\bb{x}\}$, $\{\bb{y}\}$, and $\{\bb{z}\}$ shown in Fig. \ref{fig_transform_1}(a) is a member of a family of MCMs containing two contexts $\{\bb{x},\bb{y}\}$ and $\{\bb{z}\}$, shown in Fig. \ref{fig_transform_1}(b), one of which is associated to a \emph{family} of JPDs over $\bb{x}$ and $\bb{y}$ (the dash-dotted circle in Fig. \ref{fig_transform_1}(b)) which, if marginalized, produces the same $\bP(\bb{x})$ and $\bP(\bb{y})$ in the original MCM (left-most MCM). Mathematically, the set of all JPDs over RVs $\bb{x}$ and $\bb{y}$ which, if marginalized, produce specific  marginal probability distributions $\bP(\bb{x})$ and $\bP(\bb{y})$ is denoted by $\{\bP(\bb{x},\bb{y})\} \models \bP(\bb{x}) \wedge \bP(\bb{y})$. The notion of the nested property enables us to look at one MCM as a subset of another larger MCM. The nested property, furthermore, enables one to sort MCMs in a hierarchical construct as illustrated in Fig. \ref{fig_transform_1} where moving from the left to the right corresponds to moving from lower levels of hierarchy to higher levels.
\begin{figure}[h!] 
  \centering
    \includegraphics[width=0.39\textwidth]{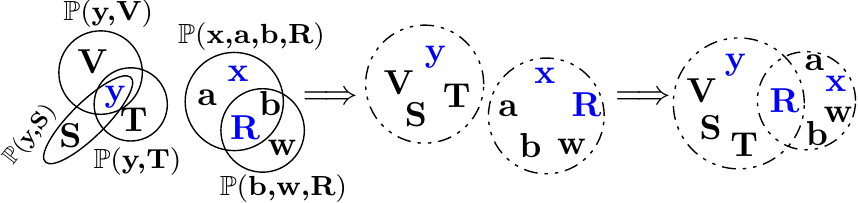}
  \caption{Transformation: Sample case.}
  \label{fig_reduction}
\end{figure}

To convey the idea, consider the case illustrated in Fig. \ref{fig_reduction}. Suppose the query of interest is $\bP(x|y,R)_\dd$. Then, one can first transform the original (left-most) MCM into the MCM shown in the middle, and subsequently into the right-most MCM. Hence, using the right-most MCM and the rule given in Fig. \ref{fig_grammar_cases}(b), one can write $\bP(x|y,R)_\dd=[\frac{\bP(x|R)-\bP(\bar{y}|R)}{\bP(y|R)}]^+ =[\frac{\bP(x|R)-1+\bP(y|R)}{\bP(y|R)}]^+$. If we had the knowledge of $\bP(y|R)$ then the expression given above would have been sufficient to derive $\bP(x|y,R)_\dd$. However, since $\bP(y|R)$ is \emph{not} known, we need to go through one more step. This is precisely due to, and  emphasizes, the fact that by working on the right-most MCM we implicitly presumed that we were equipped with more knowledge than we really had.  Using the middle MCM and the rule given in Fig. \ref{fig_grammar_cases}(a), one can conclude $\textstyle\bP(y|R)_\dd=\textstyle[\frac{\bP(y)-\bP(\bar{R})}{\bP(R)}]^+$. Altogether\footnote{This is due to the observation that for function $ f(y)=\textstyle(\frac{k+y}{y})$ when $k<0$, $\textstyle\min_{1\geq y\geq t>0}f(y)=\textstyle(\frac{k+t}{t})$.}, $\bP(x|y,R)_\dd = \big( [\frac{\bP(x|R)-1+\bP(y|R)}{\bP(y|R)}]^+\big)_{\dd}= [\frac{\bP(x|R)-1+\bP(y|R)_\dd}{\bP(y|R)_\dd}]^+$. It is worth noting that the same rule would apply if instead of the random vector $\bb{R}$ we were dealing with the random variable $\bb{a}$, i.e., to find $\bP(x|y,a)_\dd$ one could use the same expression given for $\bP(x|y,R)_\dd$ by substituting $a$ in place of $R$ in all the expressions. Arguments of this kind are made possible due to the idea of transformation which enables us to analyze the transformed MCM (e.g., the middle one in Fig. \ref{fig_reduction}) rather than the original MCM (the left-most one in Fig. \ref{fig_reduction}). Furthermore, the concept of transformation highlights a key idea: if a piece of information (i.e., an intra-contextual quantity) is irrelevant in the transformed MCM for the posed query, it must have been irrelevant in the original MCM in the first place. This statement, once again, sheds light on what intra-contextual quantities are relevant or irrelevant to derive a posed inter-contextual query on a given MCM.

%% file: Discussion.tex
\section{DISCUSSION}
\label{sec:work}
We will now discuss related work so as to build a connection between ours and previous attempts to incorporate partial probabilistic knowledge of a domain in the task of inference.

Attempting to combine Probabilistic Logic and BNs, the authors in 
\cite{andersen1990probabilistic,andersen1994bayesian} formulate the inference problem as an optimization problem subject to non-linear constraints so as to incorporate the conditional independence relations embedded in the BN. However, in our proposed framework, the issue of dealing with conditional independence relations does not arise at all, because these relations are dealt with during the derivation process of intra-contextual probabilities. 

The authors of \cite{hansen1995models} point out that one could avoid non-linear optimization when the value for a conditional probability is at least imprecisely known. For example, the constraint $\bP(a|b)=\bP(a)$, if the value for $\bP(a)$ is known either precisely or imprecisely within some interval $[\alpha,\beta]$, can be written as 
\begin{eqnarray*}
\frac{\bP(a,b)}{\bP(b)}=\bP(a) \in [\alpha,\beta] \Leftrightarrow
\left\{
\begin{array}{lc}
\bP(a,b)-\alpha \bP(b)>0,& \\
\bP(a,b)-\beta \bP(b)<0.&
\end{array}
\right.
\end{eqnarray*}
Hence, the independence $\bP(a|b)=\bP(a)$ can be formulated as a number of linear constraints. However, the main drawback of this approach is that encoding a conditional independence relation such as $\bP(\bb{x}|\bb{y},\bb{a}_1,\cdots, \bb{a}_n)=\bP(\bb{x}|\bb{y})$ requires a number of linear equations that is exponential in $n$ to be introduced into the optimization problem \cite{andersen1994bayesian}.

Drawing on the idea of Context-Specific Independence (CSI) \cite{boutilier1996context}, the authors of \cite{geiger1991advances} propose the Bayesian Multinet model which aims at taking advantage of the existing CSIs to perform inference, by modeling a single BN as multiple context-specific BNs. Translated into our multi-context setting, the Bayesian Multinet model corresponds to the case where the whole domain is modeled as a single BN, i.e., a single-context MCM, that can be decomposed into multiple BNs each being valid for a specific instantiation of some RVs in the domain.    

The authors of \cite{kiefllingtowards} point out the same concerns which led us to propose MCM, namely: (i) If unverified (in)dependencies are imposed between the variables in the domain then implausible results may arise; (ii) PGMs require one to have complete probabilistic knowledge of a domain which may not be available. Motivated by these, \cite{kiefllingtowards} gives a collection of rules to carry out inference in a domain. Broadly speaking, this work is similar to ours in spirit with the main distinction being the level of abstraction chosen to perform inference. In \cite{kiefllingtowards} inference is performed in a very local and rule-based fashion and conditional independence relations are dealt with directly which complicates the task at hand; a task which is futile when it comes to dealing with domains of many variables. In our case, by introducing the notion of context and encoding conditional independence relations within contexts we avoid having to contemplate the intra-contextual inference problem and leave this task for the corresponding context. This way, we can take advantage of the possibly rich independence structure governing the context and carry out the intra-contextual inference problem in a computationally efficient manner. 

Finally, let us discuss some interesting aspects of the proposed model.

The degree of belief is encoded mathematically in the form of a probability distribution over the variables contained within the context. Furthermore, in the process of partial belief formation (which leads to the formation of contexts) the reasoner is ignorant as to how various contexts probabilistically interact (are related), except that, some contexts may in fact share a number of variables in between and hence overlap. Later on, in the process of the derivation of the query posed to the reasoner, this ignorance manifests in the uncertainty region represented by the min/max values for the inter-contextual query of interest. In other words, if the reasoner incurs ignorance as to the (in)dependency structure governing the variables present in the domain, then later on, in the process of derivation of the posed query, the reasoner has to pay the price by merely arriving at a  \emph{probability interval} rather than a point probability as an answer to the query of interest. Yet, the knowledge of the underlying dependency structure is a fundamental knowledge whose availability to the reasoner should \emph{not} be postulated as an inevitability but as an advantaged position.

The evolutionary process of MCM does not enforce a specific gradual expansion path, for the claim of MCM is merely that \emph{any partial belief formation as to the domain can be modeled in the framework depicted by MCM}. That is, the reasoner may arrive at different MCMs, depending on the order in which the reasoner encounters different concepts and also depending on her background knowledge as to the nature of the potential connections between a collection of variables. Simply put, the order according to which the reasoner comes about knowing the concepts or propositions of the domain does matter (cf. the discussion on the order of belief formation in Sec. \ref{Sec:Gen}).

MCM enables one to carry out inference without having to commit to any unjustified or uncertain independence assumptions. In light of this, contexts symbolize the regions of the domain over which an (in)dependence structure is presumed and hence, the growth and merging of contexts indicates the formation of new (in)dependence structures over some parts of the domain which previously were unstructured. In short, MCM is meant to be invoked in circumstances where the observations and the a priori knowledge combined are not sufficient for the reasoner to form the full JPD over all of the domain variables and yet, quite crucially, the reasoner is reluctant to submit to any unjustified assumptions to compensate for such inadequacy of knowledge.

\section{CONCLUSION}
\label{sec:conclusion}
In an attempt to establish a middle ground between Bayesian Logic and Probabilistic Logic  \cite{andersen1990probabilistic,andersen1994bayesian}, on one side, and PGMs\footnote{For instance, Bayesian Networks \cite{pearl1986fusion}, Markov Networks \cite{koller2009probabilistic}, and Chain Graphs \cite{buntine1995chain}.} on the other, we proposed the Multi-Context Model to represent the state of partial knowledge regarding a domain. The generative process for the gradual construction of contradiction-free MCMs was discussed. The task of Inference for MCM was studied and, along the path, the notions of inference grammar, nestedness, and transformation were introduced. A short version of $\mathcal{I}^\ast$ without the scale-invariance property was provided in Appendix. It is worth noting that scale-invariance property can be achieved with a minor change to the last step of the proposed algorithm.

%% file: Appendix.tex
\section*{\centerline{APPENDIX}}
\subsubsection*{A-I $\mathcal{I}_{non-scale}^\ast$: A short version of $\mathcal{I}^\ast$ without scale-invariance property}
\label{A-I}
$\mathcal{I}^\ast$ aims at minimally parameterizing the information contained in an MCM so that the posed inter-contextual query can be stated as an LP with the fewest number of parameters. As pointed out earlier in Sec. \ref{Sec:Grammar}, $\mathcal{I}^\ast$ has to decide on the following: (i) what RVs have to be included in the LP, and (ii) the abstraction level required to minimally encode the information on the RVs identified in step (i) for the LP, in our case, the parametrization of the identified RVs. 

In what follows, a simple algorithm, $\mathcal{I}_{non-scale}^\ast$, is sketched which only performs (i) and ignores (ii). In other words, $\mathcal{I}_{non-scale}^\ast$  identifies the relevant RVs needed to derive the \emph{exact} lower/upper bound for the inter-contextual query, however, it does not aim at minimally encoding them into the LP\footnote{To read more on this, the reader is referred to the discussion on scale-invariance property in Sec. \ref{Sec:Grammar} and Sec. A-III of Appendix.}.  $\mathcal{I}_{non-scale}^\ast$ consists of three steps:
\begin{itemize}
\item[(1)] Identify all the RVs involved in the posed query (e.g., in $P(X|Y,z)$ these are the random vector $X$, random vector $Y$ and RV $z$).
\item[(2a)] If any two of the already identified RVs belong to two overlapping contexts, identify all the \emph{overlapping} RVs between these two contexts (e.g., in Fig. 5(b) and for the query $P(X|Y)$ for which step (1) would identify $X$ and $Y$, random vector $Z$ in the overlapping region must be identified as well).
\item[(2b)] If any two of the already identified RVs belong to two contexts connected through a chain of overlapping contexts: identify all the RVs contained in all the \emph{overlapping} regions of the chain of contexts.
\item[(3)] Parameterize only the identified RVs in steps (1), (2a), and (2b) (remove all the other RVs from the MCM---there is no need to encode the information on any other RVs not identified in steps (1), (2a), and (2b)).

\hfill $\blacksquare$
\end{itemize}

It should be noted that whether the posed query involves minimization or maximization does not affect which RVs need to be identified by $\mc I_{non-scale}^{\ast}$. Finally, It is worth noting that with a minor modification to step (3) of $\mathcal{I}_{non-scale}^\ast$, the scale-invariance property could be achieved. The modification has to do with the question of how to \emph{minimally} encode the information on each RV identified in steps (1), (2a), and (2b) of $\mathcal{I}_{non-scale}^\ast$.

\begin{figure}[h!] 
  \centering
    \includegraphics[width=0.38\textwidth]{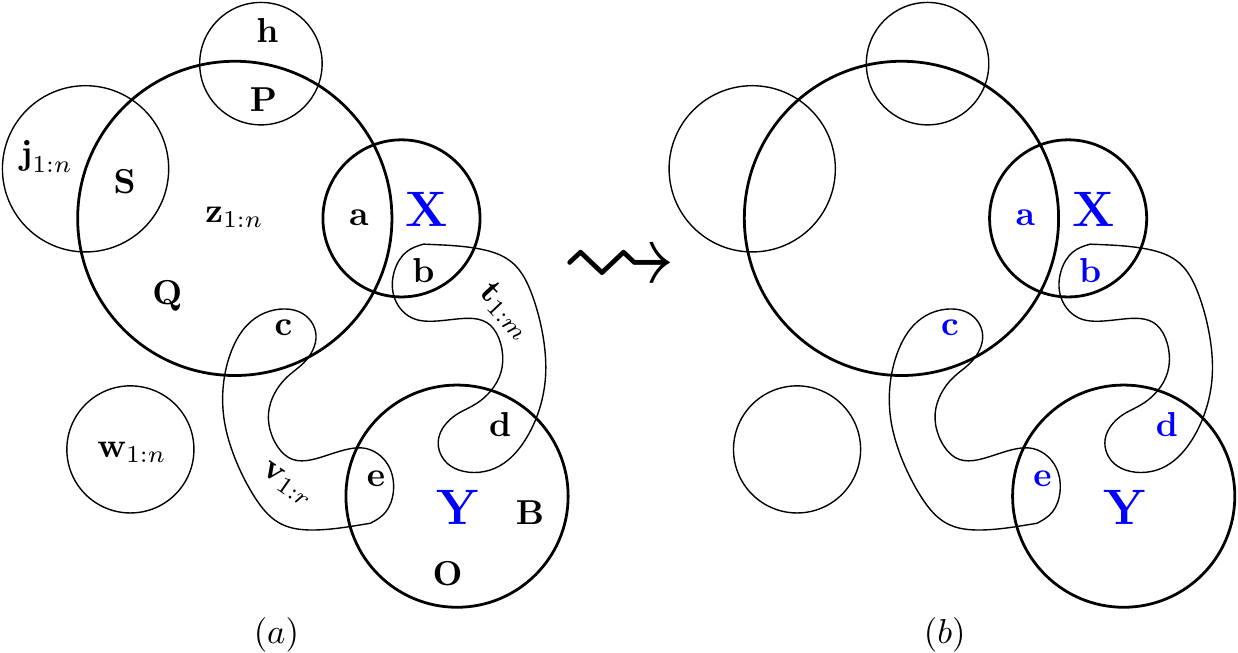}
  \caption{(a) Sample MCM. The RVs involved in the posed query are depicted in blue. (b) In Step (1) $\bb{X}$ and $\bb{Y}$ are identified; in step (2b) the RVs $\bb{b}$, $\bb{d}$ as well as $\bb{a}$, $\bb{c}$, and $\bb{e}$ are identified. According to step (3) of $\mc I_{non-scale}^{\ast}$ all of the information as to the RVs $\bb{X}$, $\bb{Y}$, $\bb{b}$, $\bb{d}$, $\bb{a}$, $\bb{c}$, and $\bb{e}$ has to be stated as an LP to derive the query.}
  \label{A_alg_sample}
\end{figure}

To demonstrate the operation of $\mc I_{non-scale}^{\ast}$ on a  more complicated MCM that involves loops, consider the following example sketched in Fig. \ref{A_alg_sample}(a). The query of interest is $\bP(X|Y)_\dd$.

Next, we are going to sketch the proof for $\mc I_{non-scale}^{\ast}$. Let us first state the claim formally and then provide the proof.

\subsubsection*{A-II Proof for $\mc I_{non-scale}^{\ast}$:}
\textbf{Lemma}: \emph{Given a posed query and an MCM, if all the information on the RVs identified in steps (1) to (2b) of $\mc I_{non-scale}^{\ast}$ is stated and then solved as an LP, the exact solution (i.e., a min or max) can be derived for the posed query; all the remaining information available in the MCM is deemed irrelevant to the derivation of the query, hence the sufficiency.}

\textbf{Proof:} Our proof is constructive. In the proof we entertain two ideas, namely (i) the idea of generative process and, particularly, that of \emph{conditioning}  also used in Sec. \ref{Sec:Gen}, and (ii) the notion we refer to as the \emph{locality of information}. Suppose that all the RVs discussed in steps (1) to (2b) of $\mc I_{non-scale}^{\ast}$ are identified. The key insight is that the information on how the remaining RVs probabilistically interact with each other is completely local in nature and, therefore, irrelevant to the derivation of the posed query. To see this, one can start off with the identified RVs and then in a gradual fashion add on\footnote{This is based on the fundamental property that a JPD can be expanded using the chain rule of probability in an arbitrary order.} the rest of the RVs (through the idea of conditioning discussed in Sec. \ref{Sec:Gen}). Quite crucially, this very process of adding the non-identified RVs to the model can be done completely in a local fashion, i.e., without imposing any constraints on how the identified RVs probabilistically interact. The mere fact that those RVs can be added into the model: (i) subsequent to the identified ones, and (ii) without inducing any sort of constraints on the identified ones, deems them irrelevant to the derivation of the query. \hfill $\blacksquare$

\subsubsection*{A-III Scale-Invariance Property: Intuition}
\begin{figure}[h!] 
  \centering
    \includegraphics[width=0.48\textwidth]{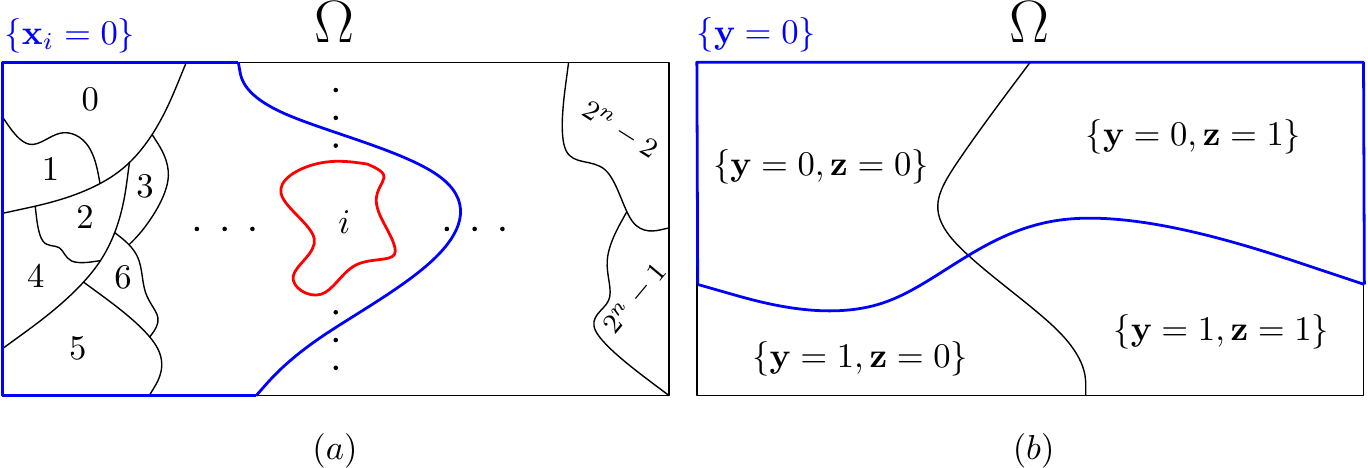}
  \caption{Sample Space: (a) Partitioning induced on $\Omega$ due to $\bb{X}=\bb{x}_{1:n}$. The blue region corresponds to the partition associated to the event $\{\bb{x}_i=0\}$ and the red one to that of $\{\bb{X}=i\}$ where $i \in \VV(\bb{X})$. (b) Partitioning induced on $\Omega$ due to RVs $\bb{y}$ and $\bb{z}$. The blue region corresponds to the partition associated to the event $\{\bb{y}=0\}$.}
\label{fig_scale_inv}
\end{figure}

Here, we will provide a proof for the example on scale-invariance property given in Sec. \ref{Sec:Grammar}. Although the proof is provided for a special query, the methodology used in the proof provides an insightful way of \emph{visualizing} an inference problem. The idea behind the proof is very simple and related to visualizing the connection of a RV to the underlying sample space using Venn diagrams. Without loss of generality, we assume that all the RVs present in the domain are binary\footnote{The generalization of the argument to non-binary RVs is straightforward.}. Random vector $\bb{X}=\bb{x}_{1:n}$ partitions the sample space $\Omega$ into $2^n$ disjoint regions each of which corresponds to a realization of $\bb{X}$. If each realization of the random vector $\bb{x}_{1:n}$ corresponds to a binary number (i.e., binary-coding the realizations), then one can conclude $\VV(\bb{X})=\{0,1,\cdots,2^n-1\}$. Let us index the partitions by their corresponding realization of $\bb{X}$. An illustrative example  of an induced partitioning of the sample space $\Omega$ due to random vector $\bb{X}=\bb{x}_{1:n}$ is depicted in Fig. \ref{fig_scale_inv}(a), and a partitioning induced by RVs $\bb{y}$ and $\bb{z}$ is sketched in Fig. \ref{fig_scale_inv}(b). We note that the mere knowledge of the distribution function of a random quantity does not provide one with the knowledge of the underlying partitions. For this particular example, since the JPD over $\bb{X},\bb{y},\bb{z}$ is not available, the knowledge of how the partitions induced by $\bb{y},\bb{z}$ (Fig. \ref{fig_scale_inv}(b)) and the ones induced by $\bb{X}$ (Fig. \ref{fig_scale_inv}(a)) interact, i.e., to what extent they overlap, remains unspecified. Therefore, since $\bP(X|y)=\frac{\bP(X,y)}{\bP(y)}$, to minimize (maximize) $\bP(X|y)$, the quantity $\bP(X,y)$ has to be minimized (maximized).  Pictorially, the minimization (maximization) of $\bP(X,y)$ corresponds to the minimization (maximization) of the overlap between the partitions corresponding to the events $\{\bb{X}=X\}$ and $\{\bb{y}=y\}$; hence, very simply, $\textstyle\bP(X,y)_\dd=[\bP(X)+\bP(y)-1]^+$ and $\textstyle\bP(X,y)_\uu=\min\{\bP(X),\bP(y)\}$. The key point, which yields the scale-invariance property, is that to derive the minimum (maximum) overlap between the partitions corresponding to the events $\{\bb{X}=X\}$ and $\{\bb{y}=y\}$ 
\emph{the information as to how the other partitions---corresponding to the other realizations of the present RVs in the model---interact with one another neither needs to be known nor to be encoded into the LP}; a fact which results in not requiring to encode the information as to the other realizations. Hence the only pieces of information that are required to be encoded and then solved as an LP are $\bP(X)$ and $\bP(y)$. The same line of reasoning could be adopted for $\bP(x_i|y)$. The idea of scale-invariance, therefore, aims to avoid the encoding of the information as to the partitions induced on $\Omega$ which are yet deemed to be irrelevant to the derivation of the posed query; hence one needs to encode solely the relevant ones into the LP.

\subsubsection*{Acknowledgement}
The authors would like to thank the anonymous reviewers for their valuable comments. 

This work was supported in part by the Natural Sciences and Engineering Research Council (NSERC) under grant RGPIN 262017 and by the Fonds Quebecois de la Recherche sur la Nature et les Technologies (FQRNT).